\documentstyle[aclap]{article}

\newtheorem{proposition}{Proposition}
\newtheorem{theorem}{Theorem}

\newcommand{\tuple}[1]{\mbox{$\langle$#1$\rangle$}}
\newcommand{\proofend}{\begin{flushright} \mbox{$\Box$} \end{flushright}}

\author{Koichi Takeda \\
    Tokyo Research Laboratory, IBM Research \\ 1623-14 Shimotsuruma,
Yamato, Kanagawa 242, Japan \\ Phone: 81-462-73-4569, 81-462-73-7413
(FAX) \\ {\tt takeda@trl.vnet.ibm.com}}
\title{Pattern-Based Context-Free Grammars for Machine Translation}

\begin{document}

\bibliographystyle{fullname}

\maketitle


\vspace{-0.5in}

\begin{abstract}

This paper proposes the use of ``pattern-based'' context-free grammars 
as a basis for building machine translation (MT) systems, which are 
now being adopted as personal tools by a broad range of users in the 
cyberspace society. We discuss major requirements for such tools,
including easy customization for diverse domains, the efficiency of 
the translation algorithm, and scalability (incremental improvement in
translation quality through user interaction), and describe how our
approach meets these requirements. 

\end{abstract}

\section{Introduction}

With the explosive growth of the World-Wide Web (WWW) as information
source, it has become routine for Internet users to access textual data 
written in foreign languages. In Japan, for example, a dozen or so
inexpensive MT tools have recently been put on the market to help PC
users understand English text in WWW home pages. 
The MT techniques employed in the tools, however, are fairly conventional. 
For reasons of affordability, their designers appear to have made no 
attempt to tackle the well-known problems in MT, such as how to ensure
the learnability of correct translations and facilitate customization.
As a result, users are forced to see the same kinds of translation errors
over and over again, except they in cases where they involve merely
adding a missing word or compound to a user dictionary, or specifying one
of several word-to-word translations as a correct choice. 

There are several alternative approaches that might eventually 
liberate us from this limitation on the usability of MT systems:

Unification-based grammar formalisms and lexical-semantics formalisms 
(see LFG \cite{kaplb82}, HPSG \cite{polls87}, and Generative 
Lexicon \cite{pust91}, for example) have been proposed to facilitate
computationally precise description of natural-language syntax and semantics. 
It is possible that, with the descriptive power of these grammars and 
lexicons, individual usages of words and phrases may be defined 
specifically enough to give correct translations.
Practical implementation of MT systems based on these formalisms, 
on the other hand, would not be possible without much more efficient 
parsing and disambiguation algorithms for these formalisms and a method 
for building a lexicon that is easy even for novices to use.

Corpus-based or example-based MT  \cite{sato90,sumita91} and 
statistical MT \cite{brown93} systems provide the easiest
customizability, since users have only to supply a collection of source
and target sentence pairs (a bilingual corpus).
Two open questions, however, have yet to be satisfactorily answered 
before we can confidently build commercial MT systems based on these
approaches:
\begin{itemize}
\item Can the system be used for various domains without showing 
      severe degradation of translation accuracy? 
\item What is the minimum number of examples (or training data) 
      required to achieve reasonable MT quality for a new domain?
\end{itemize}

TAG-based MT \cite{abeille90}\footnote{See LTAG \cite{schabes88}
(Lexicalized TAG) and STAG \cite{shies90} (Synchronized TAG) 
for each member of the TAG (Tree Adjoining Grammar) family.} 
and pattern-based translation \cite{maru93} share many important
properties for successful implementation in practical MT systems, namely:
\begin{itemize}
\item The existence of a polynomial-time parsing algorithm
\item A capability for describing a larger {\it domain of 
locality} \cite{schabes88}
\item {\it Synchronization} \cite{shies90}  
of the source and target language structures
\end{itemize}

Readers should note, however, that the parsing algorithm for TAGs
has $O(|G|n^{6})$\footnote{Where $|G|$ stands for the size of grammar G,
and $n$ is the length of an input string.} worst case time
complexity \cite{shanker87}, and that the ``patterns'' in Maruyama's
approach are merely context-free grammar (CFG) rules. Thus, it has been
a challenge to find a framework in which we can enjoy both a grammar
formalism with better descriptive power than CFG and more efficient
parsing/generation algorithms than those of TAGs.\footnote{
Lexicalized CFG, or Tree Insertion Grammar (TIG) \cite{schabesw95}, 
has been recently introduced to achieve such efficiency and lexicalization.}

In this paper, we will show that there exists a class of 
``pattern-based'' grammars that is weakly equivalent to CFG
(thus allowing the CFG parsing algorithms to be used for our grammars),
but that it facilitates description of the domain of locality.
Furthermore, we will show that our framework can be extended to 
incorporate example-based MT and a powerful learning mechanism.

\section{Pattern-Based Context-Free Grammars}

{\it Pattern-based context-free grammars} (PCFG) consists of 
a set of translation {\it patterns}. A pattern is a pair of CFG 
rules, and zero or more {\it syntactic head} and {\it link} 
constraints for nonterminal symbols. For example, the English-French
translation pattern\footnote{and its inflectional variants ---
we will discuss inflections and agreement issues later.}
\begin{quote}
NP:1 miss:V:2 NP:3 $\rightarrow$ S:2 \\
S:2 $\leftarrow$ NP:3 manquer:V:2 $\grave{a}$ NP:1
\end{quote}
essentially describes a {\it synchronized}\footnote{The meaning of the
word ``synchronized'' here is exactly the same as in
STAG \cite{shies90}. See also bilingual signs \cite{tsujii91} for a
discussion of the importance of combining the appropriate domain 
of locality and synchronization.}
pair consisting of a left-hand-side English CFG rule
(called a {\it source} rule)
\begin{quote}
NP V NP $\rightarrow$ S
\end{quote}
and a French CFG rule (called a {\it target} rule)
\begin{quote}
S $\leftarrow$ NP V $\grave{a}$ NP
\end{quote}
accompanied by the following constraints.
\begin{enumerate}
\item {\bf Head constraints:} The nonterminal symbol V in the source
rule must have the verb {\it miss} as a syntactic head. The symbol V in
the target rule must have the verb {\it manquer} as a syntactic head. The
head of symbol S in the source (target) rule is identical to the head of
symbol V in the source (target) rule as they are co-indexed.
\item {\bf Link constraints:} Nonterminal symbols in source and target
CFG rules are {\it linked} if they are given the same index ``:$i$''.
Linked nonterminal must be derived from a sequence of synchronized pairs.
Thus, the first NP (NP:1) in the source rule corresponds to the second
NP (NP:1) in the target rule, the Vs in both rules correspond to each
other, and the second NP (NP:3) in the source rule corresponds to the
first NP (NP:3) in the target rule.
\end{enumerate}
The source and target rules are called {\it CFG skeleton} of the
pattern.  The notion of a syntactic head is similar to that used in
unification grammars, although the heads in our patterns are simply
encoded as character strings rather than as complex feature structures.
A head is typically introduced\footnote{A nonterminal symbol $X$ in a
source or target CFG rule $X \leftarrow X_{1} \cdots X_{k}$ can only be
constrained to have one of the heads in the RHS $X_{1} \cdots X_{k}$. 
Thus, {\it monotonicity} of head constraints holds throughout the parsing
process.} in preterminal rules such as
\begin{quote}
leave $\rightarrow$ V V $\leftarrow$ partir
\end{quote}
where two verbs, ``leave'' and ``partir,'' are associated with the heads
of the nonterminal symbol V. This is equivalently expressed as
\begin{quote}
leave:1 $\rightarrow$ V:1 V:1 $\leftarrow$ partir:1
\end{quote}
which is physically implemented as an entry of an English-French lexicon.

A set T of translation patterns is said to {\it accept} an input $s$ iff
there is a derivation sequence Q for $s$ using the source CFG skeletons
of T, and every head constraint associated with the CFG skeletons in Q
is satisfied. Similarly, T is said to {\it translate} $s$ iff there is
a synchronized derivation sequence Q for $s$ such that T accepts $s$,
and every head and link constraint associated with the source and target
CFG skeletons in Q is satisfied.  The derivation Q then produces a
translation $t$ as the resulting sequence of terminal symbols included
in the target CFG skeletons in Q.  Translation of an input string $s$
essentially consists of the following three steps:
\begin{enumerate}
\item Parsing $s$ by using the source CFG skeletons
\item Propagating link constraints from source to target CFG
skeletons to build a target CFG derivation sequence
\item Generating $t$ from the target CFG derivation sequence
\end{enumerate}
The third step is a trivial procedure when the target CFG derivation
is obtained.

\begin{theorem}
Let T be a PCFG. Then, there exists a CFG $G_{T}$ such that for two
languages $L(T)$ and $L(G_{T})$ accepted by T and $G_{T}$, respectively,
$L(T) = L(G_{T})$ holds. That is, T accepts a sentence s iff $G_{T}$ 
accepts s. 
\end{theorem}
{\bf Proof:} We can construct a CFG $G_{T}$ as follows:
\begin{enumerate}
\item $G_{T}$ has the same set of terminal symbols as T.
\item For each nonterminal symbol X in T, $G_{T}$ includes a 
set of nonterminal symbols \{$X_{w} | w$ is either a terminal symbol 
in T or a special symbol $\epsilon$\}.
\item For each preterminal rule
\begin{quote}
$X$:i $\leftarrow$ $w_{1}$:1~$w_{2}$:2~$\ldots$~$w_{k}$:$k~(1 \leq i
\leq k)$,
\end{quote}
$G_{T}$ includes\footnote{Head constraints are trivially satisfied or
violated in preterminal rules. Hence, we assume, without loss of generality,
that no head constraint is given in preterminal rules. We also assume that 
``$X \leftarrow w$'' implies ``$X$:1 $\leftarrow$ $w$:1''.}
\begin{quote}
$Xw_{i} \leftarrow w_{1}~w_{2}~\ldots~w_{k}$ $(1 \leq i \leq k)$.
\end{quote}
If X is not co-indexed with any of $w_{i}$, $G_{T}$ includes
\begin{quote}
$X_{\epsilon} \leftarrow w_{1}~w_{2}~\ldots~w_{k}$.
\end{quote}
\item For each source CFG rule with head constraints ($h_{1}, h_{2}$,
$\ldots, h_{k}$) and indexes ($i_{1}, i_{2}, \ldots, i_{k}$),
\begin{quote}
$Y$:$i_{j}$ $\leftarrow$ 
$h_{1}$:$X_{1}$:$i_{1}$~$\ldots~h_{k}$:$X_{k}$:$i_{k}$ $(1 \leq j \leq k)$,
\end{quote}
$G_{T}$ includes 
\begin{quote}
$Yh_{j} \leftarrow Xh_{1}~Xh_{2}~\ldots~Xh_{k}$.
\end{quote}
If $Y$ is not co-indexed with any of its children, we have
\begin{quote}
$Y_{\epsilon} \leftarrow Xh_{1}~Xh_{2}~\ldots~Xh_{k}$.
\end{quote}
If $X_{j}$ has no head constraint in the above rule, $G_{T}$ includes 
a set of $(N+1)$ rules, where $Xh_{j}$ above is replaced with $X_{w}$ for 
every terminal symbol $w$ and $X_{\epsilon}$
($Yh_{j}$ will also be replaced if it is co-indexed
with $X_{j}$).\footnote{Therefore, a single rule in T can be mapped to 
as many as $(N+1)^{k}$ rules in $G_{T}$, where N is the number of terminal
symbols in T. $G_{T}$ could be exponentially larger than T.}
\end{enumerate}
Now, $L(T) \subseteq L(G_{T})$ is obvious, since $G_{T}$ can simulate the 
derivation sequence in T with corresponding rules in $G_{T}$. 
$L(G_{T}) \subseteq L(T)$ can be proven, with mathematical induction, from
the fact that every valid derivation sequence of $G_{T}$ satisfies head
constraints of corresponding rules in T. \proofend

\begin{proposition}
Let a CFG G be a set of source CFG skeletons in T. Then,
$L(T) \subseteq L(G)$.
\end{proposition}
Since a valid derivation sequence in T is always a valid derivation
sequence in G, the proof is immediate. Similarly, we have
\begin{proposition}
Let a CFG H be a subset of source CFG skeletons in T such
that a source CFG skeleton $k$ is in H iff $k$ has no head constraints
associated with it. Then, $L(H) \subseteq L(T)$.
\end{proposition}
Two CFGs G and H define the range of CFL L(T). These two CFGs can be
used to measure the ``default'' translation quality, since idioms
and collocational phrases are typically translated by patterns with
head constraints.

\begin{theorem}
Let a CFG G be a set of source CFG skeletons in T. Then,
$L(T) \subset L(G)$ is undecidable.
\end{theorem}
{\bf Proof:} The decision problem, $L(T) \subset L(G)$, of  
two CFLs such that $L(T) \subseteq L(G)$ is 
solvable iff $L(T) = L(G)$ is solvable. This includes a known undecidable
problem, $L(T) = \Sigma^{*}?$, since we can choose a grammar U
with $L(U) = \Sigma^{*}$, nullify the entire set of rules in U by
defining T to be a vacuous set \{$S$:1 $\leftarrow$ $a$:$S_{b}$:1, 
$S_{b}$:1 $\leftarrow$ b:$S_{U}$:1\} $\cup$ U ($S_{U}$ and S are start 
symbols in U and T, respectively),
and, finally, let T further include an arbitrary CFG F. 
$L(G) = \Sigma^{*}$ is obvious, since G has 
\{$S$ $\leftarrow$ $S_{b}$, $S_{b}$ $\leftarrow$ $S_{U}$\} $\cup$ U.
Now, we have $L(G) = L(T)$ iff $L(F) = \Sigma^{*}$. \proofend

Theorem 2 shows that the syntactic coverage of T is, in general,
only computable by T itself, even though T is merely a CFL. 
This may pose a serious problem when a
grammar writer wishes to know if there is a specific expression
that is only acceptable by using at least one pattern with head
constraints, for which the answer is ``no'' iff L(G) = L(T). 
One way to trivialize this problem is to let T include a pattern 
with a pair of pure CFG rules for every pattern with head constraints, 
which guarantees that L(H) = L(T) = L(G). In this case, we know that
the coverage of ``default'' patterns is always identical to L(T).

Although our ``patterns'' have no more theoretical descriptive power 
than CFG, they can provide considerably better descriptions of the
domain of locality than ordinary CFG rules. For example,
\begin{quote}
be:V:1 year:NP:2 old $\rightarrow$ VP:1 \\
VP:1 $\leftarrow$ avoir:V:1 an:NP:2
\end{quote}
can handle such NP pairs as ``one year'' and ``un an,'' and ``more than
two years'' and ``plus que deux ans,'' which would have to be covered by
a large number of plain CFG rules. TAGs, on the other hand, are
known to be ``mildly context-sensitive'' grammars, and they can capture
a broader range of syntactic dependencies, such as cross-serial
dependencies. The computational complexity of parsing for
TAGs, however, is $O(|G|n^{6})$, which is far greater than that of CFG
parsing.  Moreover, defining a new STAG rule is not as easy for the
users as just adding an entry into a dictionary, because each STAG rule
has to be specified as a pair of tree structures. Our
patterns, on the other hand, concentrate on specifying linear ordering of
source and target constituents, and can be written by the users as easily 
as\footnote{By sacrificing linguistic accuracy for the description of
syntactic structures.}
\begin{quote}
to leave * = de quitter * \\
to be year:* old = d'avoir an:*
\end{quote}
Here, the wildcard ``*'' stands for an NP by default. The
preposition ``to'' and ``de'' are
used to specify that the patterns are for VP pairs,
and ``to be'' is used to show that the phrase is the BE-verb and its
complement. A wildcard can be constrained with a head, as in ``house:*''
and ``maison:*''.  
The internal representations of these patterns are as follows:
\begin{quote}
leave:V:1 NP:2 $\rightarrow$ VP:1 \\
VP:1 $\leftarrow$ quitter:V:1 NP:2
\\[2mm]
be:V:1 year:NP:2 old $\rightarrow$ VP:1 \\
VP:1 $\leftarrow$ avoir:V:1 an:NP:2
\end{quote}
These patterns can be associated with an explicit nonterminal
symbol such as ``V:*'' or ``ADJP:*'' in addition to head constraints
(e.g., ``leave:V:*'').  By defining a few such notations, these patterns
can be successfully converted into the formal representations defined
in this section. Many of the divergences \cite{dorr93b} in source and 
target language expressions are fairly collocational, and can be
appropriately handled by using our patterns.
Note the simplicity that results from using a notation in which users
only have to specify the surface ordering of words and phrases. More
powerful grammar formalisms would generally require either a structural
description or complex feature structures. 

\section{The Translation Algorithm}

The parsing algorithm for translation patterns can be any of known CFG
parsing algorithms including CKY and Earley algorithms\footnote{Our
prototype implementation was based on the Earley algorithm, since this
does not require lexicalization of CFG rules.} At this stage,
head and link constraints are ignored. It is easy to show that the
number of target charts for a single source chart increases exponentially
if we build target charts simultaneously with source charts. 
For example, the two patterns
\begin{quote}
A:1 B:2 $\rightarrow$ B:2 B:2 $\leftarrow$ A:1 B:2, and \\ 
A:1 B:2 $\rightarrow$ B:2 A:1 $\leftarrow$ B:2 A:1
\end{quote}
will generate the following $2^{n}$ synchronized pairs of charts 
for the sequence of (n+1) nonterminal symbols $A A A \ldots A B$,
for which no effective packing of the target charts is possible.
\begin{quote}
(A (A $\ldots$ (A B))) with (A (A $\ldots$ (A B))) \\
(A (A $\ldots$ (A B))) with ((A $\ldots$ (A B)) A) \\
$\ldots$ \\
(A (A $\ldots$ (A B))) with (((B A) A) $\ldots$ A) \\
\end{quote}
Our strategy is thus to find a candidate set of source charts
in polynomial time. We therefore apply heuristic measurements to identify
the most promising patterns for generating translations. In this sense,
the entire translation algorithm is not guaranteed to run 
in polynomial time. Practically, a timeout mechanism and a process for
recovery from unsuccessful translation
(e.g., applying the idea of fitted parse \cite{jensen83} to target CFG
rules) should be incorporated into the translation algorithm. 

Some restrictions on patterns must be imposed to avoid infinitely 
many ambiguities and arbitrarily 
long translations. The following patterns are therefore not allowed:
\begin{enumerate}
\item $A \rightarrow X$ $Y \leftarrow B$
\item $A \rightarrow X$ $Y \leftarrow C_{1} \ldots B \ldots C_{k}$ \\
\end{enumerate}
if there is a cycle of synchronized derivation such that
\begin{quote}
$A \rightarrow X \ldots \rightarrow A$ and \\
$B$ (or $C_{1} \ldots B \ldots C_{k}$) $\rightarrow Y 
\ldots \rightarrow B$, \\
\end{quote}
where A, B, X, and Y are nonterminal symbols with or without head and
link constraints, and C's are either terminal or nonterminal symbols.

The basic strategy for choosing a candidate derivation sequence 
from ambiguous parses is as follows.\footnote{This strategy is
similar to that of transfer-driven MT (TDMT) \cite{furuse94}.
TDMT, however, is based on a combination of declarative/procedural
knowledge sources for MT, and no clear computational properties 
have been investigated.} A simplified view of the Earley 
algorithm \cite{earl7002}
consists of three major components, {\it predict(i), complete(i),}
and {\it scan(i)}, which are called at each position $i = 0, 1,
\ldots, n$ in an input string I = $s_{1} s_{2} \ldots s_{n}$.
Predict(i) returns a set of currently applicable CFG rules at 
position $i$. Complete(i) combines inactive charts ending at $i$
with active charts that look for the inactive charts at position $i$
to produce a new collection of active and inactive charts. Scan(i)
tries to combine inactive charts with the symbol $s_{i+1}$ at position
$i$. Complete(n) gives the set of possible parses for the input I.

Now, for every inactive chart associated with a nonterminal symbol X
for a span of \tuple{i,j} $(1 \leq i,j \leq n)$, 
there exists a set P of patterns with the source CFG skeleton, 
$\ldots \rightarrow X$. 
We can define the following ordering of patterns in P;
this gives patterns with which we can use head and link constraints
for building target charts and translations. These candidate patterns
can be arranged and associated with the chart in the complete() procedure.

\begin{enumerate}
\item Prefer a pattern $p$ with a source CFG skeleton 
$X \leftarrow X_{1} \cdots X_{k}$ over any other pattern $q$ with
the same source CFG skeleton $X \leftarrow X_{1} \cdots X_{k}$,
such that $p$ has a head constraint $h$:$X_{i}$  if $q$ has $h$:$X_{i}$ 
$(i = 1, \ldots, k)$. The pattern $p$ is said to be {\it more specific} 
than $q$. For example, p = ``leave:V:1 house:NP $\rightarrow$ VP:1'' 
is preferred to q = ``leave:V:1 NP $\rightarrow$ VP:1''.
\item Prefer a pattern $p$ with a source CFG skeleton to any
pattern $q$ that has fewer terminal symbols in the source CFG skeleton
than $p$. For example, prefer ``take:V:1 a walk'' to ``take:V:1 NP''
if these patterns give the VP charts with the same span.
\item Prefer a pattern $p$ which does not violate any head constraint
over those which violate a head constraint.
\item Prefer the shortest derivation sequence for each input substring.
A pattern for a larger domain of locality tends to give a shorter
derivation sequence.
\end{enumerate}

These preferences can be expressed as numeric values ({\it cost}) for 
patterns.\footnote{A similar preference can be defined for the target
part of each pattern,
but we found many counter-examples, where the number of nonterminal
symbols shows no specificity of the patterns, in the target part of 
English-to-Japanese translation patterns. Therefore, only the head
constraint violation in the target part is accounted for in our prototype.}
Thus, our strategy favors {\it lexicalized} (or head constrained) and
{\it collocational} patterns, which is exactly what we are going to
achieve with pattern-based MT.  Selection of patterns in the
derivation sequence accompanies the construction of a target chart.
Link constraints are propagated from source to target
derivation trees. This is basically a bottom-up procedure.

Since the number M of distinct pairs \tuple{X,w}, for a nonterminal symbol
$X$ and a subsequence $w$ of input string $s$, is bounded by $Kn^{2}$,
we can compute the {\it m-best choice} of pattern candidates for every
inactive chart in time $O(|T|Kn^{3})$ as claimed by Maruyama \cite{maru93}, 
and Schabes and Waters \cite{schabesw95}. Here, $K$ is the number of 
distinct nonterminal symbols in T, and $n$ is the size of the input string.
Note that the head constraints associated with the source CFG rules can
be incorporated in the parsing algorithm, since the number of triples
\tuple{X,w,h}, where h is a head of X, is bounded by $Kn^{3}$. We can
modify the predict(), complete(), and scan() procedures to run in 
$O(|T|Kn^{4})$ while checking the source head constraints. Construction
of the target charts, if possible, on the basis of the m best candidate
patterns for each source chart takes $O(Kn^{2}m)$ time. Here, $m$ can
be larger than $2^{n}$ if we generate every possible translation.

The reader should note critical differences between lexicalized grammar
rules (in the sense of LTAG and TIG) and translation patterns when they
are used for MT. 

Firstly, a pattern is not necessarily lexicalized. 
An economical way of organizing translation patterns is to include
non-lexicalized patterns as ``default'' translation rules.  

Secondly, lexicalization might increase the size of STAG grammars
(in particular, compositional grammar rules such as ADJP NP 
$\rightarrow$ NP) considerably when a large number of phrasal
variations (adjectives, verbs in present participle form, 
various numeric expressions, and so on) multiplied by the number of
their translations, are associated with the ADJP part.
The notion of structure sharing \cite{shanker92} may have to be 
extended from lexical to phrasal structures, as well as from
monolingual to bilingual structures.

Thirdly, a translation pattern can omit the tree structure of a
collocation, and leave it as just a sequence of terminal symbols.
The simplicity of this helps users to add patterns easily, although
precise description of syntactic dependencies is lost. 

\section{Features and Agreements}

Translation patterns can be enhanced with unification and feature
structures to give patterns additional power for describing gender, 
number, agreement, and so on.
Since the descriptive power of unification-based grammars is considerably
greater than that of CFG \cite{berw8207}, feature structures have to be
restricted to maintain the efficiency of parsing and generation algorithms.
Shieber and Schabes briefly discuss the issue \cite{shies90}. We can 
also extend translation patterns as follows:
\begin{quote}
Each nonterminal node in a pattern can be associated with a fixed-length
{\it vector} of {\it binary features}. 
\end{quote}
This will enable us to specify such syntactic dependencies as
agreement and subcategorization in patterns. Unification of binary 
features, however, is much simpler: unification of a feature-value
pair succeeds only when the pair is either \tuple{0,0} or \tuple{1,1}.
Since the feature vector has a fixed length, unification of two
feature vectors is performed in a constant time. For example, the
patterns\footnote{Again, these patterns can be mapped to a 
weakly equivalent set of CFG rules. See GPSG \cite{gazdp85} for more
details.}
\begin{quote}
V:1:+TRANS NP:2 $\rightarrow$ VP:1 VP:1 $\leftarrow$ V:1:+TRANS NP:2 \\
V:1:+INTRANS $\rightarrow$ VP:1 VP:1 $\leftarrow$ V:1:+INTRANS
\end{quote}
are unifiable with transitive and intransitive verbs, respectively.
We can also distinguish {\it local} and {\it head} features, as 
postulated in HPSG. Simplified version of verb subcategorization
is then encoded as
\begin{quote}
VP:1:+TRANS-OBJ NP:2 $\rightarrow$ VP:1:+OBJ 
VP:1:+OBJ $\leftarrow$ VP:1:+TRANS-OBJ NP:2 \\
\end{quote}
where ``-OBJ'' is a local feature for head VPs in LHSs, while ``+OBJ''
is a local feature for VPs in the RHSs. Unification of a local feature
with +OBJ succeeds since it is not {\it bound}.

Agreement on subjects (nominative NPs) and finite-form verbs (VPs, 
excluding the BE verb) is disjunctively specified as
\begin{verbatim}
NP:1:+NOMI+3RD+SG VP:2:+FIN+3SG
NP:1:+NOMI+3RD+PL VP:2:+FIN-3SG
NP:1:+NOMI-3RD    VP:2:+FIN-3SG
NP:1:+NOMI        VP:2:+FIN+PAST
\end{verbatim}
which is collectively expressed as
\begin{verbatim}
NP:1:*AGRS VP:2:*AGRV
\end{verbatim}
Here, *AGRS and *AGRV are a pair of {\it aggregate} unification
specifiers that succeeds only when one of the above combinations
of the feature values is unifiable.

Another way to extend our grammar formalism is to associate weights 
with patterns. It is then possible to rank the matching patterns 
according to a linear ordering of the weights
rather than the pairwise partial ordering of patterns described in the 
previous section. In our prototype system, each pattern has its original
weight, and according to the preference measurement described in the
previous section, a penalty is added to the weight to give the effective 
weight of the pattern in a particular context. Patterns with the least 
weight are to be chosen as the most preferred patterns.

Numeric weights for patterns are extremely useful as means of assigning
higher priorities uniformly to user-defined patterns.
Statistical training of patterns can also be incorporated to calculate
such weights systematically \cite{fuji89}.

Figure~\ref{sample} shows a sample translation of the input
``He knows me well,'' using the following patterns. 

\begin{quote}
NP:1:*AGRS VP:1:*AGRS $\rightarrow$ S:1 \\
S:1 $\leftarrow$ NP:1:*AGRS VP:1:*AGRS ... (a)
\\[1.5mm]
VP:1 ADVP:2 $\rightarrow$ VP:1 \\
VP:1 $\leftarrow$ VP:1 ADVP:2 ... (b)
\\[1.5mm]
know:VP:1:+OBJ well $\rightarrow$ VP:1 \\
VP:1 $\leftarrow$ connaitre:VP:1:+OBJ bien ... (c)
\\[1.5mm]
V:1 NP:2 $\rightarrow$ VP:1:+OBJ \\
VP:1:+OBJ $\leftarrow$ V:1 NP:2:-PRO ... (d)
\\[1.5mm]
V:1 NP:2 $\rightarrow$ VP:1:+OBJ \\
VP:1:+OBJ $\leftarrow$ NP:2:+PRO V:1 ... (e)
\end{quote}

To simplify the example, let us assume that we have the following
preterminal rules:

\begin{quote}
he $\rightarrow$ NP:+PRO+NOMI+3RD+SG \\
NP:+PRO+NOMI+3RD+SG $\leftarrow$ il ... (f)
\\[1.5mm]
me $\rightarrow$ NP:+PRO+CAUS+SG-3RD \\
NP:+PRO+CAUS+SG-3RD $\leftarrow$ me ... (g)
\\[1.5mm]
knows $\rightarrow$ V:+FIN+3SG \\
V:+FIN+3SG $\leftarrow$ sait ... (h) \\
knows $\rightarrow$ V:+FIN+3SG \\
V:+FIN+3SG $\leftarrow$ connait ... (i)
\\[1.5mm]
well $\rightarrow$ ADVP ADVP $\leftarrow$ bien  ... (j) \\
well $\rightarrow$ ADVP ADVP $\leftarrow$ beaucoup  ... (k)
\end{quote}

\begin{figure}

\baselineskip=0.60\normalbaselineskip

\begin{verbatim}

Input: He knows me well

Phase 1: Source Analysis

[0 1] He    ---> (f) NP 
   (active arc [0 1] (a) NP.VP)
[1 2] knows ---> (h) V, (i) V  
   (active arcs [1 2] (d) V.NP, 
                [1 2] (e) V.NP)
[2 3] me    ---> (g) NP 
   (inactive arcs [1 3] (d) V NP, 
                  [1 3] (e) V NP)
[1 3] knows me ---> (d), (e) VP 
   (inactive arc [0 3] (a) NP VP, 
    active arcs [1 3] (b) VP.well, 
                [1 3] (c) VP.ADVP)
[0 3] He knows me ---> (a) S
[3 4] well  ---> (j) ADVP, (k) ADVP 
   (inactive arcs [1 4] (b) VP ADVP, 
                  [1 4] (c) VP ADVP)
[1 4] knows me well ---> (b), (c) VP 
   (inactive arc [0 4] (a) NP VP)
[0 4] He knows me well ---> (a) S

Phase 2: Constraint Checking

[0 1] He    ---> (f) NP 
[1 2] knows ---> (i) V, (j) V
[2 3] me    ---> (g) NP 
[1 3] knows me ---> (e) VP
   (pattern (d) fails)
[0 3] He knows me ---> (a) S
[3 4] well  ---> (i) ADVP, (j) ADVP 
[1 4] knows me well ---> (b), (c) VP 
   (preference ordering (c), (b))
[0 4] He knows me well ---> (a) S

Phase 3: Target Generation

[0 4] He knows me well ---> (a) S
[0 1] He    ---> il
[1 4] knows me well ---> (c) VP 
      well  ---> bien
[1 3] knows me ---> (e) VP
[1 2] knows ---> connait
      (h) violates a head constraint
[2 3] me    ---> me

Translation: il me connait bien 

\end{verbatim}
\caption{Sample Translation} 
\label{sample}
\end{figure}

In the above example, the Earley-based algorithm with source
CFG rules is used in Phase 1. In Phase 2, head and link 
constraints are examined, and unification of feature structures
is performed by using the charts obtained in Phase 1.
Candidate patterns are ordered by their weights and preferences.
Finally, in Phase 3, the target charts are built to generate 
translations based on the selected patterns. 

\section{Integration of Bilingual Corpora}

Integration of translation patterns with translation examples,
or {\it bilingual corpora}, is the most important extension of
our framework. There is no discrete line between patterns
and bilingual corpora. Rather, we can view them together 
as a uniform set of translation pairs with varying degrees of
lexicalization. Sentence pairs in the corpora, however, should
not be just added as patterns, since they are often redundant,
and such additions contribute to neither acquisition
nor refinement of non-sentential patterns.

Therefore, we have been testing the integration method with the
following steps. Let T be a set of translation patterns, B be a bilingual
corpus, and \tuple{s,t} be a pair of source and target sentences.
\begin{enumerate}
\item $[${\bf Correct Translation}$]$ 
 If T can translate $s$ into $t$, do nothing.
\item $[${\bf Competitive Situation}$]$ 
 If T can translate $s$ into $t'$ $(t \neq t')$, do the following:
 \begin{enumerate}
 \item $[${\bf Lexicalization}$]$
 If there is a paired derivation sequence Q of \tuple{s,t} in T, 
 create a new pattern $p'$ for a pattern $p$ used in Q such that
 every nonterminal symbol $X$ in $p$ with no head constraint is associated
 with $h$:$X$ in $q$, where the head $h$ is instantiated in $X$ of $p$.
 Add $p'$ to T if it is not already there. Repeat the addition of such
 patterns,  and assign low weights to them until the refined sequence Q
 becomes the most likely translation of $s$. For example, add
 \begin{quote}
  leave:VP:1:+OBJ considerably:ADVP:2 $\rightarrow$ VP:1 \\
  VP:1 $\leftarrow$ laisser:VP:1:+OBJ consid\^{e}rablement:ADVP:2
 \end{quote}
 if the existing VP ADVP pattern does not give a correct translation.
 \item $[${\bf Addition of New Patterns}$]$ 
 If there is no such paired derivation sequence, 
 add specific patterns, if possible, for idioms and collocations that
 are missing in T, or add the pair \tuple{s,t} to T as a translation
 pattern. For example, add
 \begin{quote}
  leave:VP:1:+OBJ behind $\rightarrow$ VP:1 \\
  VP:1 $\leftarrow$ laisser:VP:1:+OBJ 
 \end{quote}
 if the phrase ``leave it behind'' is not correctly translated.
 \end{enumerate}
\item $[${\bf Translation Failure}$]$ If T cannot translate $s$ at all, 
 add the pair \tuple{s,t} to T as a translation pattern.
\end{enumerate}

The grammar acquisition scheme described above has not yet been 
automated, but has been manually simulated for a set of 770 
English-Japanese simple sentence pairs designed for use in MT system
evaluation, which is available from JEIDA (the Japan Electronic 
Industry Development Association) \cite{jeida95}, including:
\begin{quote}
\#100: Any question will be welcomed. \\
\#200: He kept calm in the face of great danger. \\
\#300: He is what is called ``the man in the news''. \\
\#400: Japan registered a trade deficit of \$101 million,  
reflecting the country's economic sluggishness, 
according to government figures. \\
\#500: I also went to the beach 2 weeks earlier. 
\end{quote}
At an early stage of grammar acquisition, $[${\bf Addition of 
New Patterns}$]$ was primarily used to enrich the set T of patterns,
and many sentences were unambiguously and correctly translated.
At a later stage, however, JEIDA sentences usually gave several 
translations, and $[${\bf Lexicalization}$]$ with careful assignment
of weights was the most critical task.
Although these sentences are intended to test a system's ability
to translate one basic linguistic phenomenon
in each simple sentence, the result was strong evidence for our claim.
Over 90\% of JEIDA sentences were correctly translated. Among the
failures were:
\begin{quote} 
\#95: I see some stamps on the desk .
\#171: He is for the suggestion, but I'm against it. \\
\#244: She made him an excellent wife. \\
\#660: He painted the walls and the floor white.
\end{quote}
Some (prepositional and sentential) attachment ambiguities needs to be
resolved on the basis of semantic information, and scoping of coordinated
structures would have to be determined by using not only collocational 
patterns but also some measures of balance and similarities among 
constituents.

\section{Conclusions and Future Work}

Some assumptions about patterns should be re-examined when we extend
the definition of patterns. The notion of head constraints may have to be
extended into one of a set membership constraint if we need to handle 
coordinated structures \cite{kaplm88}. Some light-verb phrases cannot be
correctly translated without ``exchanging'' several feature values between
the verb and its object. A similar problem has been found in be-verb 
phrases. 

Grammar acquisition and corpus integration are fundamental
issues, but automation of these processes \cite{wat93} is still
not complete. Development of an efficient translation algorithm,
not just an efficient parsing algorithm, will make a significant
contribution to research on synchronized grammars, including
STAGs and our PCFGs.

\section*{Acknowledgments}

Hideo Watanabe designed and implemented a prototype MT system for
pattern-based CFGs, while Shiho Ogino developed a Japanese generator
of the prototype. Their technical discussions and suggestions 
greatly helped me shape the idea of pattern-based CFGs. I would
also like to thank Taijiro Tsutsumi, Masayuki Morohashi, Hiroshi
Nomiyama, Tetsuya Nasukawa, and Naohiko Uramoto for their valuable
comments. Michael McDonald, as usual, helped me write the final 
version.  


\end{document}